%% file: paper.tex
\documentclass[runningheads]{llncs}
\usepackage[T1]{fontenc}
\usepackage{graphicx}
\usepackage{booktabs}
\usepackage{adjustbox}
\usepackage{amsmath} 
\usepackage{multirow}
\usepackage{multicol}
\usepackage[misc]{ifsym}
\usepackage{authblk}
\usepackage[a4paper, left=1in, right=1in, top=1in, bottom=1in]{geometry}

\usepackage{mwe}

\begin{document}

\title{Spanish TrOCR: Leveraging Transfer Learning for Language Adaptation}

\titlerunning{Spanish TrOCR}

\author{Valentin Laurent\thanks{Work done during the internship at Qantev.}$^{2,3}$ (\Letter) \and Filipe Lauar$^{1}$ }
\authorrunning{V. Laurent and F. Lauar}
\institute{Qantev \\ 
\and KTH Royal Institute of Technology \\
\and CentraleSupélec, université Paris-Saclay \\
\email{valentinlaure@outlook.com, filipe.lauar@qantev.com}}






\maketitle              

\begin{abstract}
This study explores the transfer learning capabilities of the TrOCR architecture to Spanish. TrOCR is a transformer-based Optical Character Recognition (OCR) model renowned for its state-of-the-art performance in English benchmarks. Inspired by Li et al.'s assertion regarding its adaptability to multilingual text recognition, we investigate two distinct approaches to adapt the model to a new language: integrating an English TrOCR encoder with a language specific decoder and train the model on this specific language, and fine-tuning the English base TrOCR model on a new language data. Due to the scarcity of publicly available datasets, we present a resource-efficient pipeline for creating OCR datasets in any language, along with a comprehensive benchmark of the different image generation methods employed with a focus on Visual Rich Documents (VRDs). Additionally, we offer a comparative analysis of the two approaches for the Spanish language, demonstrating that fine-tuning the English TrOCR on Spanish yields superior recognition than the language specific decoder for a fixed dataset size. We evaluate our model employing character and word error rate metrics on a public available printed dataset, comparing the performance against other open-source and cloud OCR spanish models. As far as we know, these resources represent the best open-source model for OCR in Spanish. The Spanish TrOCR models are publicly available on HuggingFace \cite{spanish_models} and the code to generate the dataset is available on Github \cite{github}.

\end{abstract}

\section{Introduction}

\input{content/introduction}

\section{Related Work}

\input{content/related_work}

\section{Methodology}
\input{content/method}

\section{Results}
\input{content/results}

\section{Conclusion}
\input{content/conclusion}

\begin{credits}
\subsubsection{\ackname} This study was funded by Qantev.
\end{credits}

\bibliographystyle{splncs04}
\bibliography{bibliography}
\end{document}

%% file: content/introduction.tex
Optical Character Recognition (OCR), is the process of reading an image. A computer cannot read an image because it only contains pixels. OCR is the technique that translates the text represented by those pixels into a machine-understandable format. An OCR system normally is composed of a text detection and a text recognition algorithm, where the text detection problem can be seen as an object detection task where the only object is the text and the text recognition problem can be seen as a multi-modal problem where we have the cropped image as the input and the transcribed text inside this image as the output. Most of OCR research focus on the scene text recognition problem \cite{karatzas2013icdar,karatzas2015icdar,mishra2012top,phan2013recognizing,wang2011end} and/or English datasets \cite{funsd,huang2019icdar2019,jaderberg2014synthetic,gupta2016synthetic}. In this paper, we will focus on the text recognition part of OCR. More specifically, we explore the multi language adaptation capabilities of Transformer based text recognition models applied to Visual Rich Documents (VRDs).

The end goal of an OCR system is to be able to read like a human, which means that does not matter the font, the background or the writing style, it will be able to read it. However, to evaluate the quality of an OCR model, we usually take a specific dataset, split it into two sets, train/fine-tune the model in one set and evaluate it in the other set, but both are part of the same initial dataset. This approach does not test the general reading capability of the model as it is in-sample data, despite it being split between train and test sets. In this study, we propose to evaluate the OCR model in an out-of-sample way, where we pre-train the model in a large language-specific synthetic corpus adapted to VRDs and then we evaluate this model out-of-the-box in an unseen dataset.

In 2021, Li et al. \cite{li2022trocr} released a paper presenting the TrOCR architecture, a transformer-based OCR model renowned for its exceptional performance on English benchmarks. TrOCR represents a state-of-the-art approach in optical character recognition, employing transformer architectures for robust text recognition. Unlike previous OCR models, TrOCR leverages transformers for both image interpretation and text generation, enabling more efficient and accurate recognition of textual content. Its transformer-based design allows for comprehensive language understanding, facilitating superior performance across diverse linguistic contexts. In the paper, the authors claim that TrOCR can easily be extended for multilingual text recognition with minimum efforts, by just leveraging multilingual pre-trained models on the decoder side.

Motivated by this proposition, our study endeavors to explore how we can leverage these multilingual capabilities of TrOCR by adapting the model to a new language in order to recognize text on Visual Rich Documents (VRDs). We explore it for the Spanish language by generating a 2M images dataset with data augmentations that better suit the real use cases of VRDs, such as lines in the middle of the text, bars separating the characters, and artifacts in the top and in the bottom of the image caused by a propagation error from the text detection step of OCR. We then evaluate this Spanish version of TrOCR on the XFUND \cite{xfund} dataset in Spanish without further fine-tuning and compare it to other available OCR solutions.

For the TrOCR model training on Spanish, we evaluate two different approaches. In the first one, we start from the English pre-trained checkpoint of TrOCR and we train it in the Spanish data. The idea of this approach is to have an OCR that already knows how to read in English and we want to make it learn Spanish. The second approach is to initialize the TrOCR model from a Spanish text-decoder checkpoint and train it in the Spanish data. The idea in this approach is to have an OCR that cannot read but already knows Spanish, so the focus is to learn how to read.

The contributions of this paper are summarized as follows:
\begin{itemize}
    \item We provide a pipeline to create an OCR dataset from scratch for a specific language with augmentations focused on VRDs. We open source this code in: \url{https://github.com/v-laurent/VRD-image-text-generator.git}
    \item We offer a comparison between an English OCR system fine-tuned on Spanish data and an English TrOCR encoder paired with a Spanish decoder.
    \item We open-source the small, base and large Spanish TrOCR models trained, making it publicly available at HuggingFace: \url{https://huggingface.co/qantev}.
\end{itemize}

%% file: content/related_work.tex
\subsection{Models}

Optical Character Recognition is, by default, a multi-modal problem, where we have an image as input and text as output. There are two main approaches for OCR, CTC-based models and transformer-based models. 

Shi, Bai and Yao \cite{shi2016end} proposed CRNN, an end-to-end model that uses a CNN as to encode the visual information into columns and an RNN to decode it into text. They use a CTC decoding layer to post-process the output by removing the repeated symbols and all the blanks from the labels to achieve the final prediction. Multiple open-source OCR frameworks are using this architecture as their default recognition model, such as EasyOCR \cite{easyOCR} and Paddle OCR \cite{du2020pp}.

Li et al. \cite{li2022trocr} proposed TrOCR, an end-to-end transformer model that uses an image transformer as the encoder and a text transformer as the decoder. Relying fully on the transformer architecture allows the model to be flexible on the size of the architecture and the weights initialization from pre-trained checkpoints. In the paper, they propose three variants of the model: small (total parameters=62M), base (total parameters=334M) and large (total parameters=558M) versions. This diversity enables us to strike a balance between resource efficiency and parameter richness, thus enhancing the model's capability to understand language nuances and image details. 

Within the TrOCR architecture, all variants leverage vision transformers in their encoders, each with its unique architectural components. The small version adopts the DeiT architecture \cite{DeiT}, while the base and large versions incorporate a BeiT \cite{BeiT} model. Additionally, there is variation among the decoders: the small version integrates MiniLM \cite{miniLM}, while the base and large versions use RoBERTa \cite{roberta}. These three models have been trained on a dataset comprising hundreds of millions of English samples and are publicly available on the HuggingFace platform \cite{trocr_huggingface}.

An important difference between transformer and non-transformer-based models is that for the first one, we need a huge amount of data to pre-train the model due to the lack of inductive bias in transformers \cite{dosovitskiy2020image}. Without a pre-training step on synthetic data, TrOCR performance decreases a lot and the model is not able to generalize for unseen images.

\subsection{Datasets}

Most of the available benchmarks in the literature are in English and for the problem of scene text recognition. As widely used datasets for scene text recognition we have IIIT5K \cite{mishra2012top}, SVT \cite{wang2011end}, IC13 \cite{karatzas2013icdar}, IC15 \cite{karatzas2015icdar}, SVTP \cite{phan2013recognizing}, and CT80 \cite{risnumawan2014robust}. The biggest challenge of these datasets is the curved and rotated text present in the images and the different backgrounds of the text.

For documents, we have 3 widely used datasets in the literature, 2 for printed and 1 for handwritten data. For printed data, we have the SROI (Scanned Receipts OCR and Information Extraction) \cite{huang2019icdar2019} and the FUNSD (Form Understanding in Noisy Scanned Documents) \cite{funsd} datasets. These datasets contain scanned English documents of receipts and forms, that despite being real images, cannot capture the diversity of documents present in our world. For handwritten data, we have the IAM dataset containing 82,227 English words produced by 400 different writers.

The only publicly available multi-language OCR dataset on printed documents is the XFUND (Multilingual Form Understanding) dataset \cite{xfund}, which contains visually rich documents annotated in 7 different languages. However, for each language, we have only a few thousand examples, which is not enough for training an OCR system.

As the OCR models, especially the transformer ones, need millions of data in the pre-training step \cite{li2022trocr}, synthetic datasets such as MJSynth \cite{jaderberg2014synthetic} and SynthText \cite{gupta2016synthetic} are commonly used. These two datasets together have a total of 16M image-text pairs, but the problem is that they contain only English words.

To overcome the problem of having only English words, there are open source tools to generate synthetic data such as TRDG (Text Recognition Data Generator), that can generate image-text pairs for every language as it relies on the text font. The tool also allows you to generate the images with multiple augmentation techniques such as rotation, gaussian blurring, dilation, erosion and others. However, these data augmentation methods are more general computer vision techniques and do not take into account the specificities of VRDs.

%% file: content/method.tex
For an effective OCR system, a robust architecture and a comprehensive dataset are essential. Especially, the dataset should be extensive and representative enough to enable the model to learn how to read accurately in a specific language. In this section, we will explore how to generate a synthetic dataset in any language in order to train our model for the VRD problem.

\subsection{VRDs image-text generation dataset}

Developing a dataset of labeled text images needs the availability of an existing text corpus from which the images can be derived. Numerous such corpora are accessible in the public domain \cite{elgspanishtextcorpus2021,cardellino2016spanish}. However, to get a more diverse dataset that we have control over the content, we opted to extract our own corpus by scraping some Spanish Wikipedia pages. This approach also allows us to be free from existing datasets, thereby enabling us to tackle any other language. We also ensured a uniform distribution in terms of text length to prevent biasing the dataset. Throughout our experiments, we noticed that a corpus consisting of 2,000,000 sentences gives good enough generalization capabilities to the model.

To generate text images for our dataset, we explored the use of deep learning models, specifically pre-trained generative models such as ScrabbleGAN \cite{fogel2020scrabblegan} and diffusion models, such as  Diffusion-Handwriting-Generation \cite{diffusionHandwritingGeneration}.  
However, a significant challenge lies in producing accurate images that genuinely mirror the intended content. This task is made even more difficult by a persistent issue known as mode collapse. This phenomenon occurs when the generative model learns only a limited subset of the probability distribution. Consequently, the model starts generating repetitive and unrepresentative samples. This issue becomes particularly pronounced when dealing with longer sentences, leading to outputs that may not fully represent the intended diversity and complexity of the data adapted to VRDs.

The most naive approach is to artificially create text on a blank image using a specific font style. While this method may seem straightforward, it presents multiple limitations, particularly in terms of diversity and realism. These limitations become even more pronounced when we consider the complex layouts, varied fonts, presence of artifacts and intricate designs often found in VRDs. Therefore, our focus is on developing a robust tool that can effectively handle the intricacies of these visually rich documents.

To address these challenges and better represent the complexities of VRDs, we first introduced an element of randomness into some critical parameters such as font style and size, color, and padding. In our experiments, we included 20 different Spanish fonts and each image is generated based on the input text and the randomly picked parameters of font style, size, color and padding.

A key aspect of our approach was the in-depth investigation we conducted into real-world VRDs. Through this study, we were able to identify and simulate essential artifacts that are commonly found in such documents that usually downgrade the performance of the OCR. 

\begin{figure}[h]
    \centering
    \includegraphics[width=0.5\linewidth]{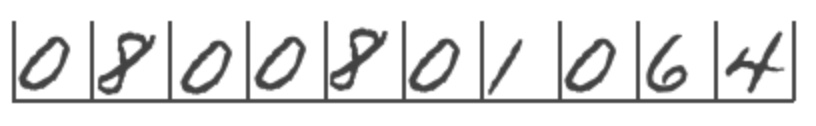}
    \caption{Boxes.}
    \label{fig:boxes}
\end{figure}

\begin{figure}[h]
    \centering
    \includegraphics[width=0.5\linewidth]{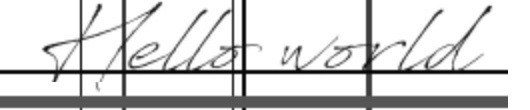}
    \caption{Random horizontal and/or vertical lines.}
    \label{fig:bars}
\end{figure}

Two artifacts intrinsically related to VRDs are the inclusion of boxes (Figure \ref{fig:boxes}) to write the information inside a form, and random vertical and/or horizontal lines \ref{fig:bars}) crossing the text. For the boxes, we want the OCR to understand that they are part of the template so it shouldn't be predicted as part of the output text. For the lines, the idea is both to present possible real-life situations to the model and also to decrease the overfitting by randomly adding these lines.

\begin{figure}[h]
    \centering
    \includegraphics[width=0.5\linewidth]{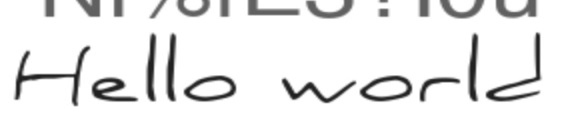}
    \caption{Cropped text.}
    \label{fig:cropped_text}
\end{figure}

\noindent Another artifact that we observed on real-life VRDs OCR applications, is the presence of text coming from the lines above or below due to a propagation error of the text detection algorithm. We observed that sometimes, especially on handwritten text, part of the text in the lines above and/or below is still present after cropping the detected text (Figure \ref{fig:cropped_text}). Therefore, we also include this artifact in our dataset so the OCR can learn how to deal with it.

\begin{figure}
    \centering
    \includegraphics[width=\linewidth]{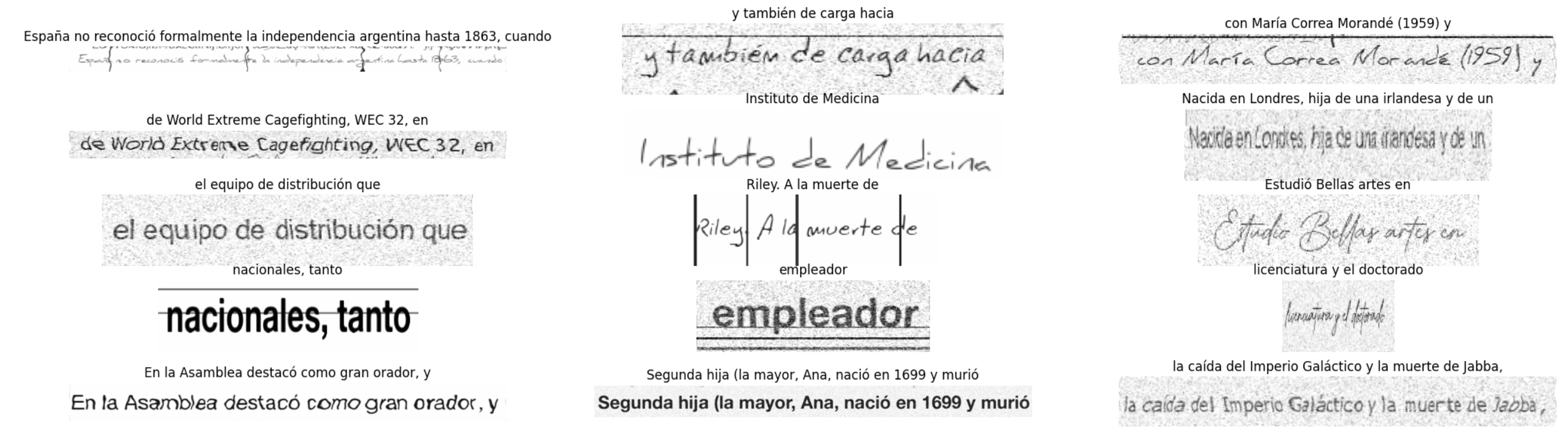}
    \caption{Few samples of the generated dataset, after the use of data augmentation. }
    \label{fig:dataset}
\end{figure}

\noindent Furthermore, we incorporated various data augmentation techniques to enhance dataset diversity and model robustness. These techniques, which mirror variations commonly found in real-world data, include random rotation, noise addition, color inversion, gaussian blur, elastic transformation, and resizing. By applying these transformations with varying probabilities, our models are exposed to a wide range of scenarios during training. This exposure improves their generalization capabilities and ensures robust performance across real-world applications for VRDs. Figure \ref{fig:dataset} provides a small subset of this generated dataset.

\noindent Given that the TrOCR model is built upon attention layers, it has a fixed capacity to comprehend context. Consequently, when dealing with lengthy texts and broader contexts, the model can become overwhelmed. To ensure that our dataset aligns with the capabilities of the TrOCR, it is crucial to establish the maximum text length that the model can effectively handle. To determine this limit, we assessed the English checkpoint of the TrOCR using English sentences of varying lengths. 

\begin{figure}
    \centering
    \includegraphics[width=\linewidth]{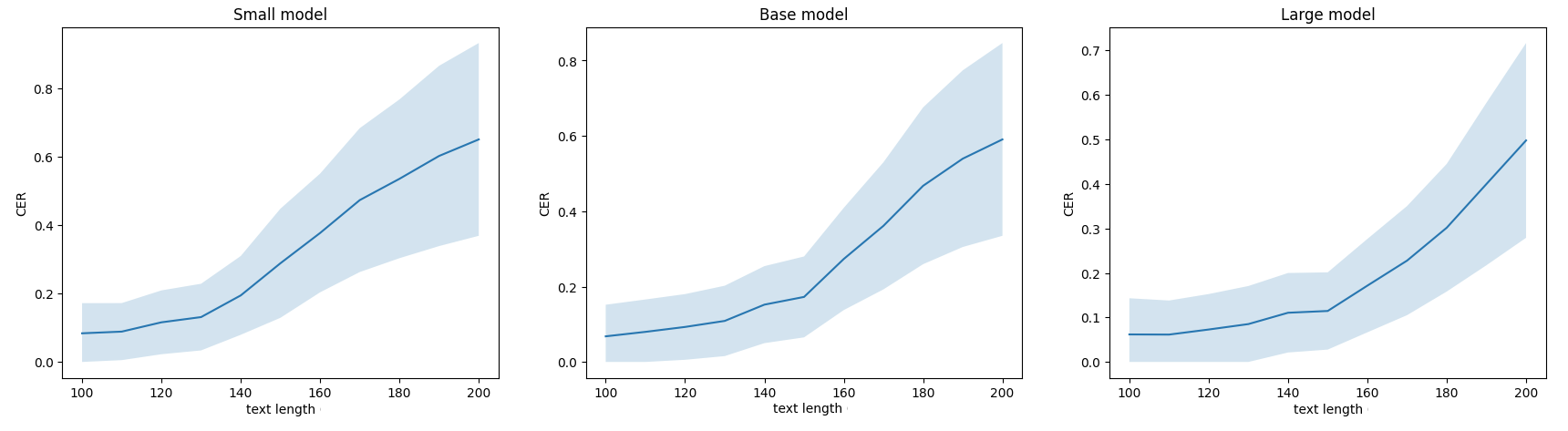}
    \caption{: CER values achieved by the small, base and large versions of the English checkpoint of the TrOCR. Each data point represents the mean value over more than a hundred images of varying sentence lengths (measured in the number of characters). The error values were computed under the assumption that the CER distribution for each point follows a Gaussian distribution.}
    \label{fig:text_length}
\end{figure}

The evaluation results are depicted in Figure \ref{fig:text_length}, illustrating the CER values achieved by the model relative to text length. Notably, the findings reveal that text length can have a substantial influence on the performance of TrOCR models. In light of this observation, we decided to restrict our dataset to texts with a maximum of 120 characters. This roughly corresponds to a maximum of 15 words, which we deemed an appropriate threshold to maintain the model's performance while ensuring the dataset's compatibility with the TrOCR's constraints.

\noindent This experiment was conducted under the assumption that the English checkpoint of TrOCR may reach its limits in terms of maximum context size. However, it is conceivable that the model was trained on a dataset where sentences longer than 120 characters were underrepresented or absent. Consequently, the model could potentially achieve higher accuracy on such sentences if it were trained on them. 

\subsection{Models}

To train the OCR model, we tried two different approaches. The first approach involves taking the English checkpoints in their three versions and fine-tuning them using the VRDs text generator to create a specialized Spanish dataset for OCR. However, it is crucial to note that this is only feasible if the initial tokenizer of the TrOCR can handle all the characters within the corpus. Therefore, this approach is only suitable for Latin languages, on which the tokenizer has been fitted. Essentially, we want the model that already knows how to read to be able to understand Spanish.

The second approach is based on the idea that the main component of the TrOCR responsible for language understanding is the text decoder. Therefore, the idea is to pair the encoder of the TrOCR with a decoder-like structure trained on Spanish data. The crucial components facilitating this merge are the cross-attention layers, which are responsible for establishing the connection between the encoder and the decoder. In our approach, we opted to link the output of the encoder directly to the input of the cross-attention layers of the decoder. Initially, the weights of these layers are randomly initialized, and subsequent training is required to optimize their performance. Essentially, this model knows Spanish and needs to learn how to read.

The primary goal of these two approaches is to compare both methods. To ensure a fair comparison, we need the models to have a similar number of parameters. To achieve this, we chose to create a small and base model with a Spanish decoder, using the MiniLM \cite{miniLM} and RoBERTa BASE \cite{roberta} Spanish versions, two models widely used for Spanish NLP tasks. These models not only share comparable architectures with the English TrOCR versions but also possess similar numbers of trainable parameters, 48M for the small model (DeiT+MiniLM) and 239M for the base model (BeiT+RoBERTa).

To train the models, we employed a single A100 80GB GPU for all training sessions. Table \ref{table:lr} displays the varying batch sizes and learning rates used for each model.

\begin{table}[h]
\centering
\caption{Learning rate used for training the various models, correlated with their respective batch sizes.}
\begin{tabular}{c|c|c}
\hline
& Batch size & Learning rate \\
\hline
Small TrOCR & 128 & $1.0 \times 10^{-5}$ \\
Base TrOCR & 64 & $7.5 \times 10^{-6}$ \\
Large TrOCR & 32 & $5.0 \times 10^{-6}$ \\
Deit+MiniLM & 128 & $1.0 \times 10^{-5}$ \\
Beit+RoBERTa & 64 & $7.5 \times 10^{-6}$ \\
\hline
\end{tabular}
\label{table:lr}
\end{table}

\noindent We trained the model for 2 epochs using the ADAM optimizer \cite{kingma2017adam} with the default parameters. We employed a step learning rate scheduler to dynamically adjust the learning rate throughout the training epochs, dividing it by two at the end of each epoch.

During our experiments, we noticed that generating images on the fly during training is faster than reading images from a folder. Consequently, we conducted the training without storing any images, which eliminates the need for extensive storage resources. To ensure consistency between the different training experiments, we set a random seed in the beginning so all the random augmentations are the same for every run. This approach not only streamlines the training process but also contributes to a more resource-efficient workflow.

The models are evaluated using the Character Error Rate (CER) and Word Error Rate (WER) metrics. These metrics measure the minimum number of insertions, deletions, and substitutions required to transition from the model's predicted text to the ground truth. To mitigate bias related to text length, a normalization factor is applied, based on the length of the ground truth.
\\
\[
    \text{CER} = \frac{I + S + D}{N} = \frac{I + S + D}{S + D + C} \quad,
\] 
    
\noindent where:

\begin{itemize}
\item $S$ is the number of character substitutions.
\item $D$ is the number of character deletions.
\item $I$ is the number of character insertions.
\item $C$ is the number of correct characters.
\item $N$ is the number of characters in the ground truth.
\end{itemize}

\[
    \text{WER} = \frac{I + S + D}{N} = \frac{I + S + D}{S + D + C} \quad,
\] 
    
\noindent where:
    
\begin{itemize}
\item $S$ is the number of word substitutions.
\item $D$ is the number of word deletions.
\item $I$ is the number of word insertions.
\item $C$ is the number of correct words.
\item $N$ is the number of words in the ground truth.
\end{itemize}

\noindent It is essential to highlight that our evaluation methodology involves computing the mean CER and WER across samples in the test dataset. Alternatively, a common approach is to calculate the weighted average based on sample length. This helps to reduce length-related bias and typically yields smaller metric values, but can be subject to misinterpretation as it does not reflect the mean value of the metric across the entire test dataset. Hence, we opted not to employ this approach and instead computed the unweighted average of CER/WER.

%% file: content/results.tex
In this section, we evaluate the data augmentation techniques applied to the model training against a fixed test set. Additionally, we evaluate the trained models on the XFUND Spanish dataset and compare their performance with two other existing Spanish OCR models.

\subsection{Image generation augmentation benchmarking}

To fully leverage the dataset creation pipeline, it is essential to benchmark all data augmentation methods. Table \ref{table:results_augmented_dataset} presents the CER and WER results for the selected OCR models under various training methods and data augmentation scenarios. The test set has 10k generated images with random augmentations. In the creation of the test set, we made sure that the input text was not present in the training set. This benchmark offers clear insights into the impact of the different augmentation techniques on model performance.

\begin{table}
    \centering
    \caption{Effect of the training method and data augmentation on the CER/WER for TrOCR small, base and large version, DeiT-MiniLM and BeiT-RoBERTa model. Comprehensive Augmentation: every data augmentation method previously discussed (except the use of the handwritten generation model). No augmentation: no data augmentation applied. No Elastic Deformation: use of all data augmentation methods previously discussed except Elastic Deformation. No artifacts: use of all data augmentation methods previously discussed except the addition of artifacts. With handwritten Generation Model: every data augmentation method previously discussed in addition to the use of the handwritten generation model ScrabbleGAN \cite{fogel2020scrabblegan} and the Diffusion-Handwriting-Generation \cite{diffusionHandwritingGeneration}.}
    \label{table:results_augmented_dataset}
    \begin{adjustbox}{width=\textwidth}
    \begin{tabular}{c c c c c c}
        \hline
        \multirow{3}{*}{Model} & \multicolumn{5}{c}{Data Augmentation Method} \\
        \cline{2-6}
         & Comprehensive & No & No Elastic & No & With HW \\
         & Augmentation & Augmentation & Deformation & Artifacts & Generation Model\\
        \hline
        Small TrOCR & 0.0184 / \textbf{0.0811} & 0.0341 / 0.0854 & \textbf{0.0178} / 0.0831 & 0.0249 / 0.0844 & 0.0452 / 0.0919 \\
        Base TrOCR & \textbf{0.0071} / \textbf{0.0387} & 0.1579 / 0.0479 & 0.0098 / 0.0416 & 0.0121 / 0.0430 & 0.0251 / 0.0518 \\
        Large TrOCR & \textbf{0.0063} / \textbf{0.0313} & 0.0125 / 0.0433 & 0.0075 / 0.0322 & 0.0095 / 0.0399 & 0.0227 / 0.0475 \\
        DeiT + MiniLM & 0.1448 / \textbf{0.5347} & 0.1629 / 0.5645 & \textbf{0.1428} / 0.5378 & 0.1574 / 0.5436 & 0.1605 / 0.6128 \\
        BeiT+RoBERTa & \textbf{0.1126} / \textbf{0.3245} & 0.1262 / 0.3378 & 0.1206 / 0.3270 & 0.1219 / 0.3401 & 0.1485 / 0.3782 \\
        \hline
    \end{tabular}
    \end{adjustbox}
\end{table}

\subsubsection{English fine-tuning vs. Spanish-decoder fine-tuning}

For the chosen set of hyper-parameters and architectures, the English version of TrOCR fine-tuned on the Spanish dataset performed significantly better than the Spanish decoder model consistently across all experiments. Also, we observed expected results when comparing the size of the models, as bigger the model, the better the results.

\subsubsection{Comprehensive vs. No Augmentation}

Incorporating all augmentation techniques generally leads to improvements in both CER and WER across all models. Indeed, no model reaches better generalization capabilities without the application of data augmentation, highlighting its importance in enhancing model performance.

\subsubsection{Elastic deformation vs. No elastic deformation}

The impact of elastic deformation as a data augmentation method varies among OCR models. The small TrOCR and Deit + MiniLM models exhibit better CER performance without elastic deformation, while the base and large TrOCR versions and BeiT+RoBERTa models benefit from it. This discrepancy may be attributed to the limited capacity of the small TrOCR and Deit + MiniLM models to handle complex cases. Interestingly, in terms of WER, these models perform better with elastic deformation, suggesting that they tend to either fully miss or fully understand a word, with the elastic deformation method accentuating this behavior.

\subsubsection{Artifacts vs. No artifacts}

Including artifacts from the training data consistently improves model performance, with reduced CER and WER values. In real-world scenarios, this data augmentation method partially addresses the issue of texts being too close together, causing text detection methods to fail in creating proper bounding boxes.

\subsubsection{Handwritten generation model: with vs. without}

The inclusion of patches generated from the handwritten generation model negatively impacts all models, with higher CER and WER values when using handwritten generated images compared to without. This issue may be attributed to the generation model's hallucinations, which result in incorrect labeling in the dataset, even when focusing only on short sentences.

\subsection{XFUND Spanish dataset}

To evaluate the capacity of our model in a real-world scenario, we used the XFUND Spanish dataset. XFUND \cite{xfund} is a multi-language dataset that contains 7 different languages. We collected the Spanish data of this dataset which contains 11,449 images-text pairs in the training set and 3,413 in the test set. It is important to mention that the XFUND has multiple problems on the annotations, we will better explore this later in this section.

To assess the real capacity of an OCR model, we believe that it should be tested in an out-of-sample dataset. Therefore, we use the pre-trained models from the synthetically generated VRD dataset and we evaluate them directly in the test set of Spanish XFUND without further fine-tuning in the training set.

We compare our results against two other OCR solutions available in the market. We compare to the well-known open-source OCR library EasyOCR-spanish, which is a CNN-LSTM based model \cite{easyOCR}. Additionally, we also compare with the Azure OCR cloud API, renowned for its efficiency. The comparison results on the XFUND dataset are presented in Table \ref{table:other_models}.

\begin{table}[h]
    \centering
    \caption{CER and WER comparison between our models and external models, such as EasyOCR and Azure OCR. As expected, Azure OCR achieved the best metrics, but it is only 2\% ahead of the large version of TrOCR in terms of CER. EasyOCR does not achieve the same results as our models, highlighting the efficiency of our method.}
    \begin{tabular}{c c c}
        \hline
        & CER & WER \\
        \hline
        Small TrOCR & 0.1059 & 0.2545 \\
        Base TrOCR & 0.0732 & 0.2028 \\
        Large TrOCR & \textbf{0.0632} & \textbf{0.1817} \\
        DeiT + MiniLM & 0.4403 & 1.1655 \\
        BeiT + RoBERTa & 0.3679 & 0.7657 \\
        \hline
        EasyOCR & 0.1916 & 0.3353 \\
        Azure OCR & \textbf{0.0429} & \textbf{0.1254} \\
        \hline
    \end{tabular}
    \label{table:other_models}
\end{table}

\noindent The fine-tuned versions of the English TrOCR outperformed the DeiT+MiniLM and BeiT+RoBERTa models in terms of CER and WER. This highlights that, for the chosen set of hyper-parameters and model architectures, fine-tuning the TrOCR models is the optimal approach for achieving cross-language adaptability. As expected, the Azure OCR version performs the best on this dataset.

\begin{figure}[h]
    \centering
    \includegraphics[width=1\linewidth]{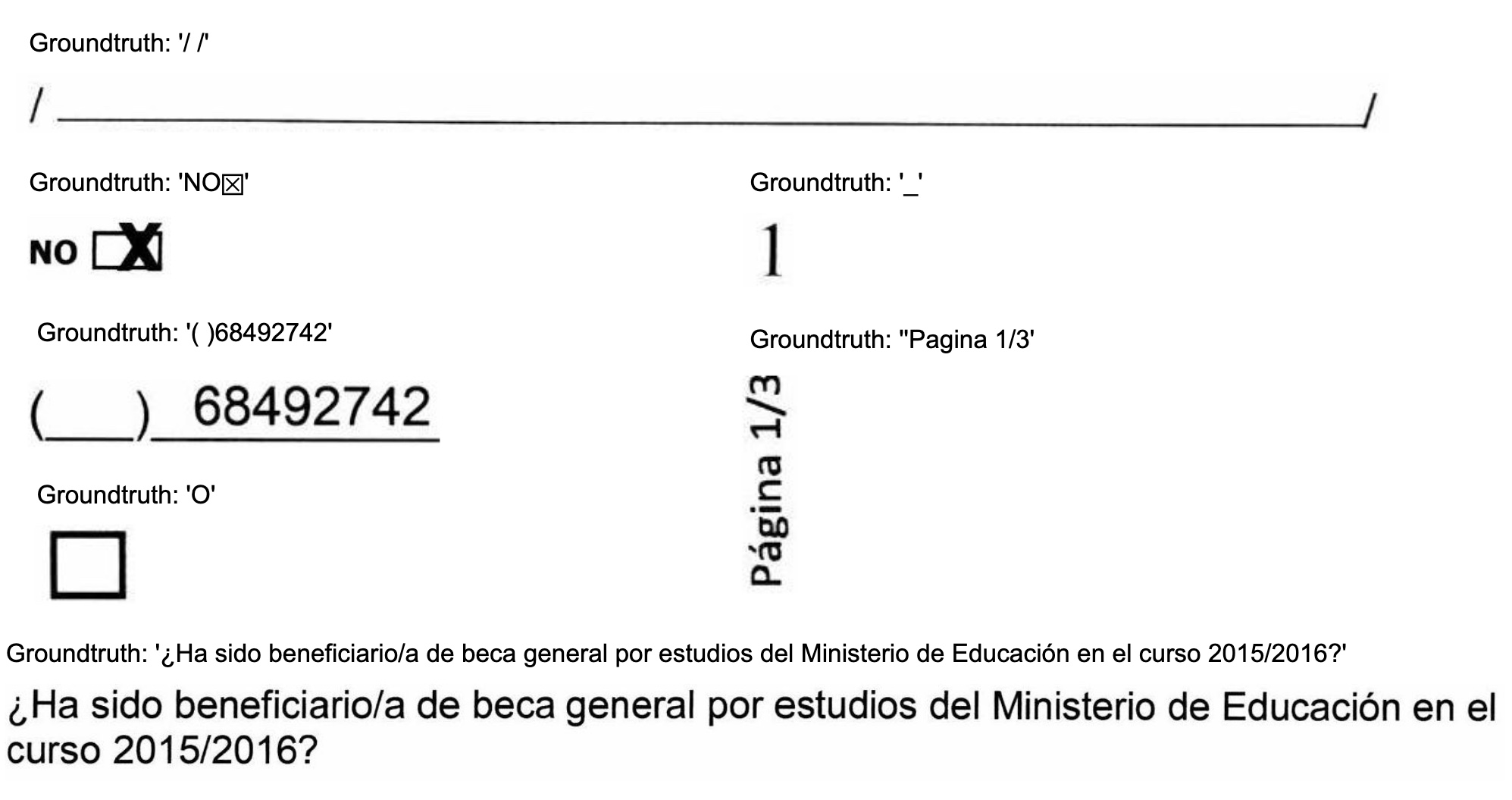}
    \caption{Example of mislabeled images on XFUND: the text orientation may be vertical, and the ground truth may incorporate non-ASCII characters, lack accent marks, or contain inaccuracies..}
    \label{fig:xfund_mislabel}
\end{figure}

Manually analyzing the mistakes made by the models, we estimate that there is a mislabeling in around 3\% of the XFUND Spanish dataset. From what we analyzed, some accents are missing in the annotations along with blank spaces that are usually predicted by the models and even some typos. We also found some images that contain more than 1 line of text, while the models are trained to recognize only a single line. Some of these mislabeling can be seen in Figure \ref{fig:xfund_mislabel}.

\subsection{Limitations and future research}

This model was not trained on handwritten data. We added handwritten fonts in the dataset generation but they do not capture the nuances of different written styles that people may have, so we do not suggest using this model for handwritten recognition without further fine-tuning.

As the English version of TrOCR, this model was trained to recognize single-line text. If you add multi-line text images, the model may hallucinate. Differently from the English TrOCR version, we did not add vertical text, so our model is not able to recognize those.

After fine-tuning the models in Spanish, they lost their capacity to predict English text when there is a similar word in Spanish. For future research, we would like to study how to make TrOCR learn a new language without losing the first one and also to include multiple languages in the same model.

For future work, we would also like to focus more on handwritten recognition and non-Latin languages using a language-specific decoder.

%% file: content/conclusion.tex
In this study we explored the multi language adaptation capabilities of TrOCR. Facing the challenge of lacking a sufficiently robust dataset for model training, specially for Visual Rich Documents, we propose a language agnostic method to synthetically generate an enormous dataset of image-text pairs to train an OCR model. This method takes into account specificities of VRDs such as boxes between character, horizontal and vertical lines and artifacts coming from the text decoder propagation error.

We explore two training approaches, fine-tuning and English checkpoint of TrOCR into Spanish and substituing TrOCR's decoder to a pre-trained Spanish one and fine-tuning it. Both approaches were fine-tuned with your synthetic Spanish dataset and we evaluated them in an out-of-sample manner on the XFUND Spanish dataset without further fine-tuning. We analyse the results based on the character error rate (CER) and word error rate (WER), considering them as the most suitable metrics for OCR evaluation.

Our findings indicate that the English TrOCR model, fine-tuned in Spanish, effectively captures information from images compared to models utilizing a mix of an English encoder and a Spanish decoder. Notably, models with an English encoder and a Spanish decoder failed to achieve metrics comparable to the fine-tuned English one. Consequently, we conclude that, for the set of hyper-parameters and architectures chosen, fine-tuning the TrOCR English version in Spanish was better than modifying the decoder to a Spanish version and then fine-tuning.

As a result, we have introduced a method for training a TrOCR in any language that can achieve high performance, comparable to professional OCR models on VRDs. This could have a significant impact on the industry, particularly considering the scarcity of publicly available OCR for languages other than English. We open-source the code to generate the dataset and the trained models.